\documentclass[letterpaper]{article} % DO NOT CHANGE THIS
\usepackage{aaai2027} % DO NOT CHANGE THIS

\usepackage[hyphens]{url} % DO NOT CHANGE THIS
\usepackage{graphicx} % DO NOT CHANGE THIS
\usepackage{natbib} % DO NOT CHANGE THIS AND DO NOT ADD ANY OPTIONS TO IT
\usepackage{caption} % DO NOT CHANGE THIS AND DO NOT ADD ANY OPTIONS TO IT
\usepackage{algorithm}
\usepackage{algorithmic}

\usepackage{newfloat}
\usepackage{listings}
\DeclareCaptionStyle{ruled}{
    labelfont=normalfont,
    labelsep=colon,
    strut=off
} % DO NOT CHANGE THIS

\floatstyle{ruled}
\newfloat{listing}{tb}{lst}{}
\floatname{listing}{Listing}

\usepackage{booktabs}
\usepackage{multirow}

\usepackage{amsmath}
\usepackage{amssymb}

\title{HYDRA: Hyperbolic Dynamic Representation Architecture for Kolmogorov-Arnold Networks}

\author{
Zhao Su\textsuperscript{\rm 1},
Yuxin Xia\textsuperscript{\rm 1},
Haoran Li\textsuperscript{\rm 2},
Jun Shen\textsuperscript{\rm 3},
Qi Zhu\textsuperscript{\rm 4},\\
Qingguo Zhou\textsuperscript{\rm 1},
Binbin Yong\textsuperscript{\rm 1}
\thanks{Corresponding author: Binbin Yong (yongbb@lzu.edu.cn)}
}

\affiliations{
\textsuperscript{\rm 1}School of Information Science and Engineering, Lanzhou University\\
\textsuperscript{\rm 2}Department of Data Science and AI, Monash University\\
\textsuperscript{\rm 3}School of Computing and Information Technology, University of Wollongong\\
\textsuperscript{\rm 4}College of Artificial Intelligence, Nanjing University of Aeronautics and Astronautics
}

\begin{document}

\maketitle

\begin{abstract}
Kolmogorov-Arnold Networks (KANs) enhance nonlinear function approximation by replacing scalar weights with learnable univariate functions. However, assigning an independent function to every connection results in substantial parameter redundancy, limiting their scalability and efficiency. To reduce this redundancy, we introduce  \textbf{HY}perbolic \textbf{D}ynamic \textbf{R}epresentation \textbf{A}rchitecture (HYDRA), a parameter-efficient hyperbolic extension of KAN that combines spline-based functional learning with representations in the Poincaré ball. HYDRA maps vector-valued inputs into a bounded hyperbolic latent space, performs KAN-style updates in tangent space, and employs a low-rank prototype block to share functional transformations across hidden dimensions. The resulting hyperbolic representations provide a structured radial coordinate for interpretation, while radius control improves training stability by preventing boundary saturation. Extensive experiments across eight benchmark datasets demonstrate that HYDRA consistently achieves competitive or superior predictive performance while improving parameter efficiency and representation interpretability.
\end{abstract}

% Links section - only shown in camera-ready version
\ifdefined\aaaianonymous
% NOTE: For anonymous submissions, do not include links that could reveal your identity
\else
%     \link{Code}{TODO}
%     \link{Datasets}{TODO}
%     \link{Extended version}{TODO}
\fi

\section{Introduction}

Many supervised learning problems with vector-valued inputs require expressive nonlinear transformations while keeping the learned model compact and inspectable. Multilayer perceptrons (MLPs) provide flexible approximation but hide feature transformations inside dense scalar weights. Kolmogorov-Arnold Networks (KANs) replace scalar weights with learnable univariate functions on edges \cite{liu2025kan}, improving local functional expressiveness but making a dense hidden-to-hidden block scale quadratically with hidden width and linearly with the number of spline bases.

The parameter-efficiency issue is closely related to how hidden representations are organized. Standard KANs operate in Euclidean spaces, where achieving stronger separation or richer representations often requires additional hidden dimensions or spline structures, increasing the parameter count. By organizing representations in a hyperbolic latent space, the model can encode variations more compactly without substantially enlarging the KAN architecture. Meanwhile, the KAN computation is performed in the tangent space, preserving the simplicity and low-parameter nature of Euclidean spline operations. Hyperbolic representation learning provides a natural alternative because distances and volumes grow differently from Euclidean spaces \cite{nickel2017poincare,ganea2018hyperbolic}. However, naive hyperbolic modeling creates a new problem: near the  Poincaré-ball boundary, distances, tangent coordinates, and gradients are amplified, so a model may separate data by drifting outward rather than by learning stable functional structure.

We propose \textbf{HY}perbolic \textbf{D}ynamic \textbf{R}epresentation \textbf{A}rchitecture (HYDRA), a radius-constrained low-rank neural network architecture that integrates hyperbolic representation learning with KAN-based function approximation. The key idea is to decouple representation scale from local functional modeling. Specifically, the Poincaré radius serves as a compact coordinate for encoding latent magnitude variations, while KAN splines capture local functional responses in the tangent space. HYDRA first maps normalized inputs into a bounded Poincaré ball, performs KAN-style residual updates in tangent coordinates, compresses these updates through a low-rank prototype bottleneck, and explicitly regulates the latent radius to prevent uncontrolled geometric expansion. Rather than assuming hierarchical structure in the input domain, HYDRA leverages hyperbolic geometry as an efficient and controllable representation space, while preserving the low-parameter and computationally efficient characteristics of KAN computation.

This design leads to the following contributions.
\begin{itemize}
    \item We propose HYDRA, a hyperbolic functional learning architecture composed of multiple HYDRA blocks, which perform spline-based KAN updates in tangent space while maintaining bounded Poincaré representations.
    \item HYDRA  introduces a low-rank prototype KAN update that reduces the dominant hidden-to-hidden parameter complexity from $O(d^2K)$ to $O(dr+r^2K)$.
    \item HYDRA incorporates a radius-control mechanism to constrain hyperbolic representations and mitigate unstable near-boundary effects.
    \item Experiments on eight datasets demonstrate that HYDRA achieves competitive or superior predictive performance while using fewer parameters than existing approaches.
\end{itemize}

\section{Related Work}

\subsection{Kolmogorov-Arnold Networks}

Kolmogorov-Arnold Networks (KANs) have renewed interest in neural architectures where nonlinear transformations are represented as explicit functions rather than implicitly encoded through dense scalar weights. In the original formulation, each edge carries a learnable univariate function, typically parameterized by splines, providing a more interpretable alternative to conventional multilayer perceptrons \cite{liu2025kan}. This formulation has motivated subsequent studies on KANs for scientific machine learning and broader neural architectures \cite{liu2024kan2,somvanshi2024kansurvey}.

Despite their interpretability advantages, dense KAN layers suffer from rapidly increasing functional parameters because each input-output connection requires an independent function. Recent variants address this limitation by modifying functional bases or introducing parameter-sharing mechanisms. Chebyshev KAN replaces spline functions with polynomial bases \cite{ss2024chebyshevkan}, Wavelet KAN introduces wavelet-based representations \cite{bozorgasl2024wavkan}, while FastKAN, radial basis function KAN, and parameter-reduced KAN variants further improve efficiency through alternative parameterizations \cite{li2024rbfkan,ta2025prkan}. These methods reduce functional representation costs while maintaining explicit function learning. However, existing KAN variants mainly focus on edge-function parameterization, leaving the role of latent representation geometry largely unexplored. HYDRA extends this line of research by coupling KAN-based functional updates with structured hyperbolic representations.

\subsection{Hyperbolic Representation Learning}

Hyperbolic representation learning studies negatively curved spaces for efficient representation of complex structures. Early studies introduced Poincaré and Lorentz embeddings, demonstrating the effectiveness of hyperbolic spaces for hierarchical representation learning \cite{nickel2017poincare,nickel2018lorentz}. These ideas were later extended to neural computation through hyperbolic entailment regions, hyperbolic neural networks, and Riemannian optimization methods \cite{ganea2018entailment,ganea2018hyperbolic,becigneul2019riemannian}.

Recent research has expanded hyperbolic learning beyond embedding problems toward complete neural architectures, including fully hyperbolic networks, hyperbolic graph models, attention mechanisms, Transformers, supervised representation learning, and residual architectures \cite{shimizu2021hyperbolicnnplusplus,chen2022fullyhyperbolic,yang2023hyperbolicrevisiting,yang2024hypformer,nock2024hyperbolicsupervised,sinha2024hypstructure,li2024lipschitzhnn,he2025lresnet}. More recently, hyperbolic geometry has also been explored for large-model adaptation and foundation-model learning \cite{yang2024hyperbolicfinetuning,he2025hyperbolicfoundation}. However, existing hyperbolic architectures mainly focus on geometric representation learning or neural computation in curved spaces, while the interaction between hyperbolic representations and explicit functional learning remains less explored. HYDRA explores this direction by performing KAN-based functional updates in tangent spaces associated with bounded hyperbolic representations.

\subsection{Interpretable Neural Networks}

Interpretability research aims to understand how neural models transform inputs into predictions. Generalized additive models and their neural extensions improve transparency by explicitly modeling feature-wise effects and interactions \cite{yang2021gaminet,agarwal2021nam,chang2022nodegam}. Meanwhile, SHAP values provide a widely adopted model-agnostic framework for estimating feature contributions \cite{lundberg2017shap}. KANs further introduce intrinsic interpretability by representing nonlinear transformations as learnable univariate functions, enabling direct analysis of feature-response relationships.

However, functional interpretability alone may not fully characterize models with structured latent spaces. HYDRA complements functional analysis by examining hyperbolic representation dynamics to provide additional insights into model behavior.

\begin{figure*}[!t]
    \centering
    \includegraphics[width=1\textwidth]{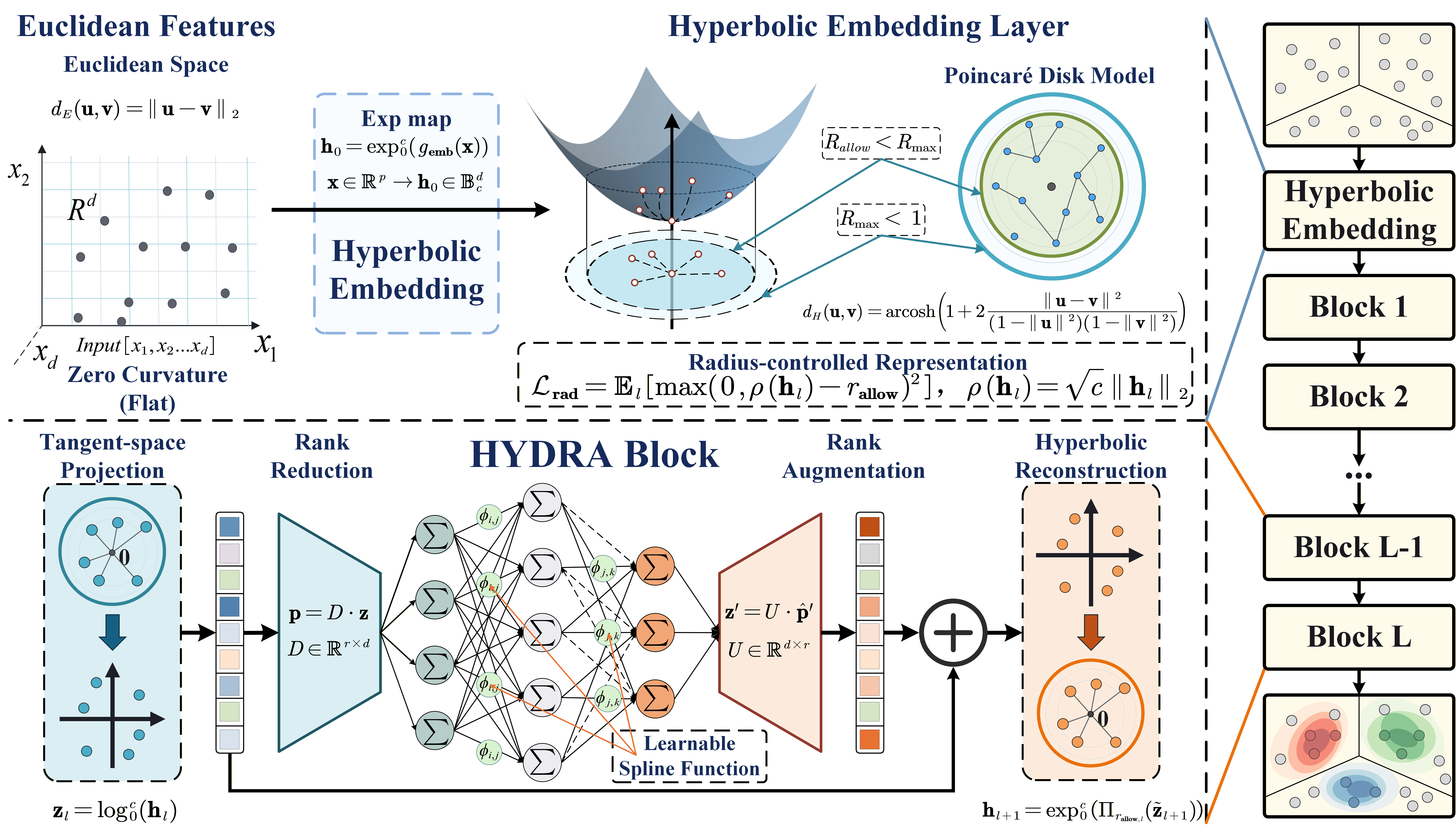}
    \caption{Overview of HYDRA. The architecture maps vector-valued features into a bounded hyperbolic representation, applies KAN-style spline updates through a low-rank prototype space, and uses radius control to stabilize geometric learning while preserving interpretable spline and radial diagnostics. All blocks share the same architecture.}
    \label{fig:hkan_method}
\end{figure*}

\section{Method}

\subsection{Problem Setup and Architecture}

We consider supervised learning with vector-valued inputs  $\mathcal{D}=\{(\mathbf{x}_i,y_i)\}_{i=1}^{n}$, where $\mathbf{x}_i\in\mathbb{R}^p$ is a normalized input and $y_i$ is either a  continuous target or a binary label. The goal is to learn a predictor  $f_\theta(\mathbf{x})$ that is accurate, parameter-efficient, and inspectable. Dense KAN layers replace scalar weights with learnable univariate functions, but a hidden-to-hidden KAN block assigns one function to each coordinate pair. This creates a dominant $O(d^2K)$ spline cost for hidden width $d$ and $K$ spline bases.

HYDRA addresses this cost by combining three operations. First, it represents hidden states in a bounded  Poincaré ball, so latent scale can be encoded through radius as well as direction. Second, it performs KAN-style functional learning in the tangent space, where ordinary one-dimensional spline bases remain available. Third, it compresses the tangent update through a learned prototype space before mapping the state back to the hyperbolic manifold. Figure~\ref{fig:hkan_method} summarizes this computation.

For layer $l$, HYDRA maps the current hyperbolic state to the tangent space, applies a low-rank spline update, and reconstructs a bounded hyperbolic state:
\begin{equation}
\begin{aligned}
\mathbf{z}_{l} &= \log_0^c(\mathbf{h}_{l}),\\
\mathbf{p}_{l} &= W_{\downarrow}\mathbf{z}_{l},\\
\mathbf{s}_{l} &= \Phi_{l}(\mathbf{p}_{l}),\\
\tilde{\mathbf{z}}_{l+1}
&=\mathbf{z}_{l}+\alpha_{l}W_{\uparrow}\mathbf{s}_{l},\\
\mathbf{h}_{l+1}
&=\Pi_{r_{l+1}}\!\left(\exp_0^c(\tilde{\mathbf{z}}_{l+1})\right),
\end{aligned}
\end{equation}
where $\Phi_l$ is a spline block in prototype coordinates, $W_{\downarrow}$ and $W_{\uparrow}$ define the bottleneck, $\alpha_l$ is a learned residual scale, and $\Pi_{r_{l+1}}$ enforces the layer radius budget. This formulation keeps local function learning Euclidean while making the hidden trajectory geometrically constrained and measurable.

\subsection{Hyperbolic Embedding and Tangent-Space Updates}

HYDRA represents hidden states on the  Poincaré ball  $\mathcal{B}_c^d=\{\mathbf{h}\in\mathbb{R}^d:c\|\mathbf{h}\|_2^2<1\}$ with curvature parameter $c>0$. The normalized input is first mapped to a tangent vector $\mathbf{u}_0=g_{\mathrm{emb}}(\mathbf{x})$. The initial hidden state is then formed by a bounded exponential map
\begin{equation}
    \mathbf{h}_0=\exp_0^c(\tilde{\mathbf{u}}_0),
    \qquad
    \|\mathbf{h}_0\|_c\le r_{\mathrm{emb}},
\end{equation}
where $r_{\mathrm{emb}}$ is the embedding radius budget. The embedding map $g_{\mathrm{emb}}$ is linear by default and can be replaced by a spline-based  input KAN when stronger input response functions are needed.

The exponential and logarithmic maps at the origin are
\begin{equation}
\begin{aligned}
\exp_0^c(\mathbf{v})
&=
\tanh(\sqrt{c}\|\mathbf{v}\|_2)
\frac{\mathbf{v}}{\sqrt{c}\|\mathbf{v}\|_2},\\
\log_0^c(\mathbf{h})
&=
\operatorname{artanh}(\sqrt{c}\|\mathbf{h}\|_2)
\frac{\mathbf{h}}{\sqrt{c}\|\mathbf{h}\|_2}.
\end{aligned}
\end{equation}
They are interpreted by continuity at the origin. These maps define a fixed coordinate chart for the KAN update. The model therefore avoids designing coordinate-wise spline functions directly on the manifold, while still returning to a hyperbolic state after each block.

Given $\mathbf{z}_l=\log_0^c(\mathbf{h}_l)$, a tangent-space KAN update has the residual form
\begin{equation}
    \mathbf{z}_{l+1}
    =
    \mathbf{z}_{l}
    +\alpha_l\Delta_l(\mathbf{z}_{l}).
\end{equation}
For a dense KAN layer, the $i$th coordinate is
\begin{equation}
\Delta_i(\mathbf{z})=\sum_{j=1}^{d}\phi_{ij}(z_j),
\qquad
\phi_{ij}(t)=\sum_{m=1}^{K}a_{ijm}B_m(t),
\end{equation}
where $B_m$ denotes spline bases and $a_{ijm}$ denotes learned coefficients. This form is expressive and interpretable, as each edge has an explicit response curve, that leading to every input-output coordinate pair owns a separate spline. HYDRA keeps the same spline principle but moves the costly functional operator into a lower-dimensional prototype space.

\newcommand{\vp}[2]{#1\,{\scriptsize(#2)}}
\newcommand{\vpm}[2]{#1\,{\scriptsize(\textbf{#2})}}

\begin{table*}[!t]
\centering

\scriptsize
\setlength{\tabcolsep}{3.2pt}
\resizebox{\textwidth}{!}{%
\begin{tabular}{lcccccccc}
\toprule
& \multicolumn{4}{c}{Regression (RMSE $\downarrow$)}
& \multicolumn{4}{c}{Classification (Accuracy $\uparrow$)} \\
\cmidrule(lr){2-5}\cmidrule(l){6-9}
Model & CCPP & Energy & Parkinsons & Real Estate & Heart & Ionosphere & Phoneme & QSAR \\
\midrule
MLP & \vp{3.779}{7.1k} & \vp{1.329}{1.6k} & \vp{5.304}{2.4k}
& \vp{\underline{6.814}}{1.8k} & \vp{0.889}{1.4k} & \vp{0.886}{1.8k}
& \vp{0.850}{1.8k} & \vp{0.882}{2.4k} \\
KAN & \vp{\underline{3.668}}{7.0k} & \vp{\underline{0.741}}{1.5k}
& \vp{4.424}{2.4k} & \vp{8.016}{1.8k} & \vp{0.833}{1.4k}
& \vp{0.843}{1.4k} & \vp{0.863}{1.8k} & \vp{0.829}{2.4k} \\
\midrule
HGCN & \vp{4.162}{7.1k} & \vp{2.420}{\textbf{1.0k}} & \vp{7.733}{2.3k}
& \vp{7.720}{1.8k} & \vp{\textbf{0.944}}{1.4k} & \vp{0.857}{1.3k}
& \vp{0.831}{1.8k} & \vp{0.877}{2.3k} \\
HNN & \vp{4.049}{7.2k} & \vp{2.273}{\textbf{1.0k}} & \vp{6.643}{2.4k}
& \vp{7.549}{1.8k} & \vp{\textbf{0.944}}{1.4k} & \vp{0.886}{1.8k}
& \vp{0.832}{1.8k} & \vp{\underline{0.891}}{2.4k} \\
GAMI-Net & \vp{3.855}{\textbf{4.8k}} & \vp{1.416}{\underline{1.1k}} & \vp{11.464}{2.4k}
& \vp{7.100}{1.8k} & \vp{\underline{0.926}}{1.7k} & \vp{\underline{0.943}}{1.7k}
& \vp{0.836}{1.8k} & \vp{0.858}{1.9k} \\
NAM & \vp{4.089}{7.1k} & \vp{3.002}{1.4k} & \vp{9.486}{2.4k}
& \vp{8.891}{1.8k} & \vp{0.778}{\underline{1.3k}} & \vp{0.800}{\underline{1.1k}}
& \vp{0.778}{1.8k} & \vp{0.796}{2.0k} \\
NODE-GAM & \vp{4.081}{14.4k} & \vp{1.138}{3.3k} & \vp{7.237}{8.1k}
& \vp{7.251}{2.7k} & \vp{0.870}{2.6k} & \vp{0.900}{3.7k}
& \vp{0.821}{6.3k} & \vp{0.853}{7.0k} \\
\midrule
FastKAN & \vp{4.303}{\underline{6.2k}} & \vp{2.077}{1.4k} & \vp{6.927}{2.8k}
& \vp{8.394}{2.0k} & \vp{0.870}{2.3k} & \vp{0.857}{1.3k}
& \vp{0.865}{1.2k} & \vp{0.839}{\underline{1.6k}} \\
ChebyKAN & \vp{3.842}{9.5k} & \vp{1.023}{1.7k} & \vp{\underline{4.078}}{2.1k}
& \vp{7.487}{\underline{1.2k}} & \vp{0.889}{1.5k} & \vp{0.857}{1.2k}
& \vp{\underline{0.885}}{\textbf{0.9k}} & \vp{0.834}{1.9k} \\
Wav-KAN & \vp{3.688}{8.7k} & \vp{2.047}{1.9k} & \vp{4.533}{1.8k}
& \vp{9.053}{\underline{1.2k}} & \vp{0.778}{1.4k} & \vp{0.871}{1.2k}
& \vp{0.880}{\underline{1.1k}} & \vp{0.863}{2.0k} \\
\midrule
\textbf{HYDRA (Ours)} & \vp{\textbf{3.604}}{\textbf{4.8k}} & \vp{\textbf{0.706}}{\underline{1.1k}}
& \vp{\textbf{3.534}}{\textbf{1.4k}} & \vp{\textbf{6.769}}{\textbf{1.0k}}
& \vp{\textbf{0.944}}{\textbf{1.2k}} & \vp{\textbf{0.971}}{\textbf{0.9k}}
& \vp{\textbf{0.885}}{\textbf{0.9k}} & \vp{\textbf{0.900}}{\textbf{1.5k}} \\
\bottomrule
\end{tabular}%
}
\caption{Primary metric and trainable parameters (in parentheses) on eight
benchmarks. Best and second-best predictive
metrics are bold and underlined; within parentheses, the smallest and
second-smallest parameter counts are bold and underlined. For detailed specifications and configuration, refer to Appendix B.}
\label{tab:main_results_with_params}
\end{table*}

\subsection{Low-Rank Prototype Functional Learning}

The low-rank block starts from the observation that hyperbolic radius can carry part of the latent scale variation. The tangent update need not always use all $d$ hidden directions independently. HYDRA therefore learns an $r$-dimensional prototype representation, applies the KAN operator there, and lifts the result  back to the hidden dimension:
\begin{equation}
    \mathbf{p}_{l}
    =
    W_{\downarrow}\mathbf{z}_{l},
    \qquad
    \Delta_l(\mathbf{z}_{l})
    =
    W_{\uparrow}\Phi_l(\mathbf{p}_{l}),
\end{equation}
where
\begin{equation}
W_{\downarrow}\in\mathbb{R}^{r\times d},
\qquad
W_{\uparrow}\in\mathbb{R}^{d\times r},
\qquad
r\ll d.
\end{equation}
The spline block $\Phi_l$ is applied before the up-projection, so the update is a nonlinear functional operator whose learned spline interactions are shared through the  prototype coordinates.

A full hidden-to-hidden KAN block contains approximately
\begin{equation}
    P_{\mathrm{full}}\approx d^2(K+1)
\end{equation}
parameters, whereas the prototype block uses
\begin{equation}
    P_{\mathrm{lr}}
    =
    2dr+r^2(K+1).
\end{equation}
The approximate compression ratio is
\begin{equation}
    \frac{P_{\mathrm{lr}}}{P_{\mathrm{full}}}
    \approx
    \frac{2}{K+1}\frac{r}{d}
    +
    \left(\frac{r}{d}\right)^2.
\end{equation}

The low-rank prototype design reduces the functional parameter cost from $O(d^2K)$ to $O(dr+r^2K)$, where the compression ratio is mainly determined by the rank ratio $r/d$.  Thus the dominant spline cost depends on $r$ rather than $d$. A small rank is useful when the tangent update has low effective dimension, while a larger rank can be selected when the task requires more independent interactions.

\subsection{Radius Control Mechanism}

Hyperbolic representations can become unstable near the boundary of the Poincaré ball. In that region, distances and tangent coordinates are amplified, so an unconstrained model may reduce the loss by pushing samples outward instead of learning smoother tangent-space functions. HYDRA controls this behavior with two mechanisms. The hard projection $\Pi_{r_l}$ keeps every layer within a prescribed radius. A soft penalty discourages unnecessary outward movement before
the projection becomes active:
\begin{equation}
    \mathcal{L}_{\mathrm{rad}}
    =
    \frac{1}{L+1}\sum_{l=0}^{L}
    \left[
    \max\left(
    0,
    \rho(\mathbf{h}_{l})-r_{\mathrm{allow},l}
    \right)
    \right]^q .
\end{equation}
Here $\rho(\mathbf{h}_l)$ is the normalized radial coordinate, $r_{\mathrm{allow},l}$ is the soft threshold, and $q$ controls the penalty shape. In practice, $r_{\mathrm{allow},l}=\tau r_l$ with $0<\tau\le 1$, where $r_l$ is the hard radius budget. Projection and penalty are complementary: projection prevents boundary violations, while the penalty changes the training direction before the representation reaches the high-amplification region.

After the final HYDRA block, prediction is made from the tangent coordinate
\begin{equation}
    \mathbf{z}_{L}=\log_0^c(\mathbf{h}_{L}),
    \qquad
    \hat{y}=g_{\mathrm{out}}(\mathbf{z}_{L}).
\end{equation}
The readout $g_{\mathrm{out}}$ is linear by default. For regression, HYDRA minimizes mean squared error on normalized targets. For binary classification, it minimizes binary cross-entropy with logits. The complete objective is
\begin{equation}
    \mathcal{L}
    =
    \mathcal{L}_{\mathrm{sup}}
    +\lambda_{\mathrm{rad}}\mathcal{L}_{\mathrm{rad}}
    +\lambda_{\mathrm{sp}}\mathcal{L}_{\mathrm{sp}},
\end{equation}
where $\mathcal{L}_{\mathrm{sup}}$ is the supervised loss and $\mathcal{L}_{\mathrm{sp}}$ combines sparsity and smoothness regularization on the spline coefficients. This objective makes HYDRA a radius-constrained, low-rank functional model. Appendix~A gives the universal approximation argument and additional bounds for the radius-controlled representation.

\section{Experiments and Results}

\subsection{Experimental Setup}

We evaluate HYDRA on eight widely used tabular benchmark datasets collected from the OpenML platform~\cite{vanschoren2013openml}: CCPP, Energy Heating, Parkinsons Telemonitoring, Real Estate Valuation, Heart Statlog, Ionosphere, Phoneme, and QSAR Biodegradation. These datasets cover both regression and binary classification tasks and exhibit diverse characteristics in terms of sample size, feature dimensionality, and underlying data distributions. Such diversity provides a representative test bed for evaluating the expressive capacity, parameter efficiency, and generalization ability of HYDRA on structured tabular data. The first four datasets are formulated as regression problems, whereas the remaining four are treated as binary classification tasks.

Regression performance is evaluated using RMSE, whereas classification performance is measured using accuracy. Additional metrics are reported in Appendix~B. For all experiments, the dataset is randomly split into training, validation, and test sets with ratios of 80\%, 10\%, and 10\%, respectively, using a fixed random seed of 42.

\paragraph{Implementation details}
All experiments are conducted on a single NVIDIA L20 GPU, and all models use the same preprocessing pipeline, optimizer configuration, and evaluation protocol. Regression models predict a single scalar and are evaluated on the original target scale, while classification models output a single logit that is converted into a probability for evaluation. Beyond the task-specific loss, HYDRA introduces only two additional regularization terms: a spline regularizer that encourages smooth and compact spline functions, and a radius regularizer that constrains hidden representations within a layer-wise radius budget. Therefore, the observed performance differences mainly reflect architectural improvements rather than variations in optimization strategies, training procedures, or computational resources.

\subsection{Evaluation Criteria}

We select models by the primary metric and report trainable parameter count to measure compactness. HYDRA hyperparameters include hidden width, prototype rank $r$, spline resolution, learning rate, weight decay, radius budget, and regularization strength. The rank is treated as a compression knob: we choose the smallest rank that preserves the main metric when possible. Parameter-free geometric operations, including exponential maps, logarithmic maps, and radius projection, are not counted as trainable capacity. The parameter-efficiency claim is evaluated primarily against Euclidean KAN and MLP because they share the same dense functional or dense hidden-state modeling role; the remaining baselines are included to contextualize predictive accuracy and interpretability-oriented alternatives.

The selection rule separates predictive quality from compactness. First, we search for HYDRA configurations that are competitive on the primary metric. Second, among configurations with similar primary performance, we prefer the one with fewer trainable parameters. This avoids two misleading extremes: a large HYDRA variant that wins mainly by capacity, and a very small HYDRA variant whose parameter count is attractive but whose accuracy is no longer competitive. The reported configuration is therefore a parameter-performance trade-off rather than the largest model found during tuning. KAN and MLP are the closest Euclidean counterparts for this comparison because they represent, respectively, dense functional edges and dense hidden-state transformations.

\begin{table}[t]
\centering
\scriptsize
\setlength{\tabcolsep}{2.5pt}
\resizebox{\columnwidth}{!}{%
\begin{tabular}{llrrrr}
\toprule
Dataset & Task & $r$ & Params F/L & Ratio & Primary F/L \\
\midrule
CCPP & Reg. & 3 & 14,536/\textbf{1,880} & 0.13 & 3.947/\textbf{3.919} \\
Energy Heating & Reg. & 3 & 1,036/\textbf{690} & 0.67 & 2.067/\textbf{1.959} \\
Parkinsons TM & Reg. & 3 & 1,211/\textbf{485} & 0.40 & 6.342/\textbf{6.037} \\
Real Estate & Reg. & 6 & 1,014/\textbf{1,000} & 0.99 & 7.273/\textbf{6.769} \\
\midrule
Heart Statlog & Cls. & 1 & 12,081/\textbf{1,091 }& 0.09 & 0.889/\textbf{0.926} \\
Ionosphere & Cls. & 7 & 932/\textbf{884} & 0.95 & 0.871/\textbf{0.943} \\
Phoneme & Cls. & 4 & 1,787/\textbf{441} & 0.25 & 0.836/\textbf{0.840} \\
QSAR Biodeg. & Cls. & 1 & 5,234/\textbf{1,440} & 0.28 & 0.877/\textbf{0.882} \\
\bottomrule
\end{tabular}%
}
\caption{Low-rank prototype ablation. F and L denote full-rank and selected low-rank HYDRA. The primary metric is RMSE for regression and Accuracy for classification.}
\label{tab:lowrank_ablation}
\end{table}

\begin{figure}[!htbp]
    \centering
    \includegraphics[width=0.5\textwidth]{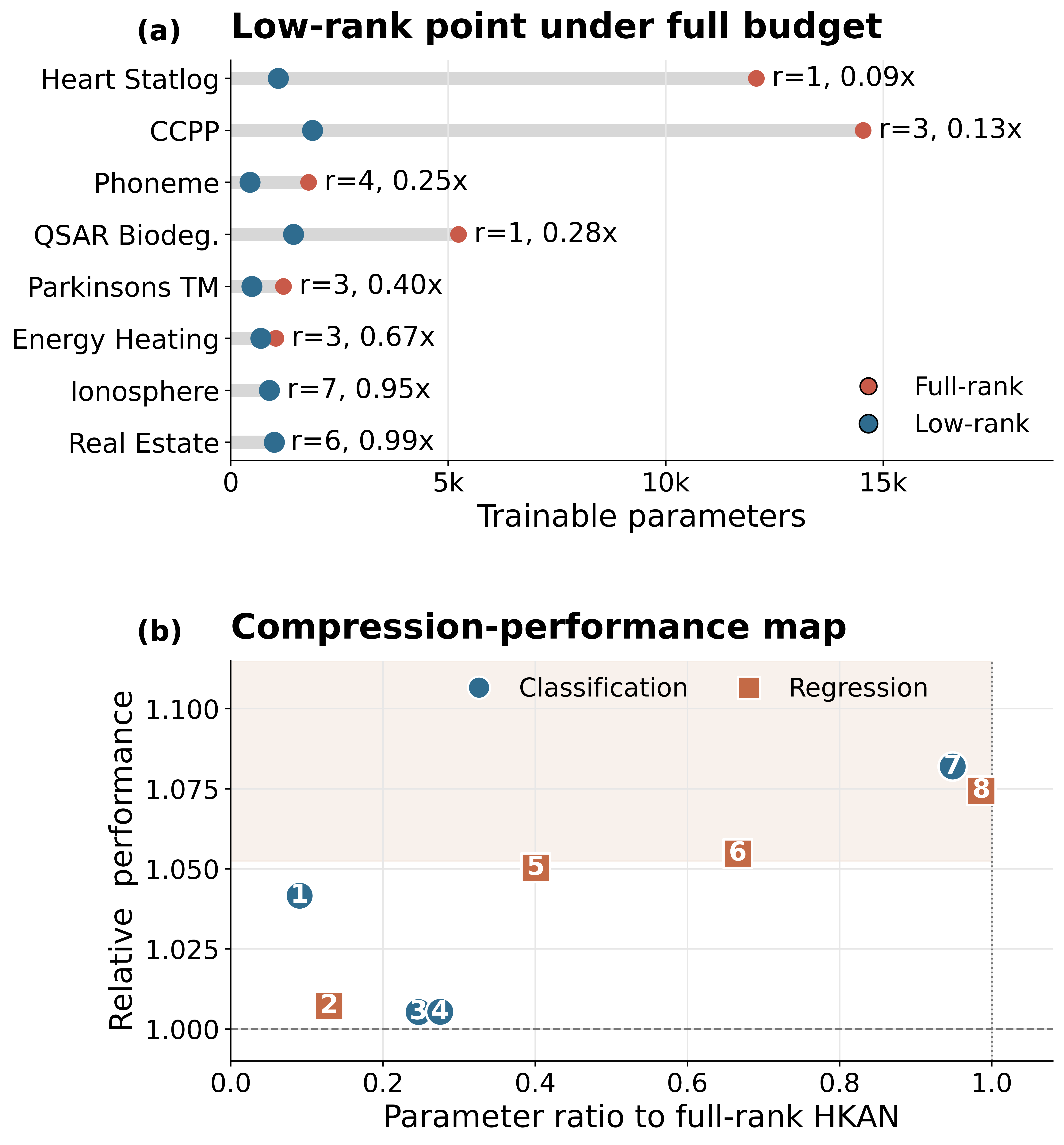}
    \caption{Low-rank ablation visualization. Selected ranks preserve the primary metric while reducing parameters relative to full-rank HYDRA. In panel~(b), numbers index the datasets after sorting by increasing parameter ratio.}
    \label{fig:lowrank_ablation}
\end{figure}

\subsection{Main Benchmark Results}

Table~\ref{tab:main_results_with_params} shows that HYDRA achieves the strongest or tied-strongest primary metric on all eight datasets. On the regression tasks, HYDRA improves over KAN and MLP on CCPP, Energy Heating, Parkinsons Telemonitoring, and Real Estate Valuation while using fewer parameters than both Euclidean counterparts. On the classification tasks, HYDRA matches the best accuracy on Heart Statlog and achieves the highest accuracy on Ionosphere, Phoneme, and QSAR Biodegradation. For example, on the Parkinsons Telemonitoring dataset, HYDRA reduces RMSE from 4.424 of KAN to 3.534 while decreasing trainable parameters from 2.4k to 1.4k, corresponding to a 20.1\% performance improvement with 41.7\% fewer parameters. Compared with MLP, HYDRA further reduces RMSE by 33.4\% under the same parameter reduction. These results support the central claim that HYDRA is not merely accurate, but accurate under a smaller trainable budget.

The benchmark mixes regression and classification because the two settings stress different aspects of the model. Regression datasets such as CCPP and Real Estate require smooth response surfaces without allocating a large number of spline edges. Classification datasets such as Ionosphere, Phoneme, and QSAR require separation while discouraging uncontrolled movement toward the  Poincaré boundary. HYDRA performs well in both regimes, suggesting that the hyperbolic coordinate is not only acting as a classifier margin and not
only as a regression smoother. Instead, it provides an internal scale coordinate that can be coupled with local spline response across objectives.

\subsection{Parameter Efficiency}

The hyperbolic maps themselves are parameter-free under fixed curvature, so HYDRA's parameter savings must come from smaller hidden widths or lower prototype ranks. Compared with KAN and MLP, HYDRA uses fewer parameters on every dataset. This matches the low-rank analysis: if the radial coordinate absorbs part of the scale-like variation, fewer prototype directions are needed for the
KAN update. The ablation below tests this interpretation against full-rank HYDRA references.

It is useful to distinguish three sources of trainable capacity. The first is the input embedding, which maps raw features into the hidden representation. The second is the hidden-to-hidden functional update. This is the dominant term for dense KAN-style models because each hidden input-output pair can own a separate spline. The third is the output readout, which is small for the widths used here. HYDRA targets the second term: the down projection, prototype spline block, and up projection replace a dense functional map with a compact bottleneck. Thus the parameter reduction is architectural rather than a post-hoc pruning effect. When the tangent update has low effective dimension, a small rank preserves the relevant directions; when many independent interactions are needed, the selected rank grows and the savings weaken.

\begin{table}[t]
\centering
\scriptsize
\setlength{\tabcolsep}{2.5pt}
\resizebox{\columnwidth}{!}{%
\begin{tabular}{llrrr}
\toprule
Dataset & Task & Primary C/U & Mean radius C/U & $\Delta$ \\
\midrule
CCPP & Reg. & \textbf{3.624}/3.690 & 0.450/0.646 & +0.066 \\
Energy Heating & Reg. & \textbf{1.193}/1.384 & 0.659/0.989 & +0.191 \\
Parkinsons TM & Reg. & \textbf{5.723}/5.883 & 0.503/0.974 & +0.160 \\
Real Estate & Reg. & \textbf{6.769}/6.884 & 0.634/0.979 & +0.115 \\
\midrule
Heart Statlog & Cls. & \textbf{0.944}/0.926 & 0.596/0.999 & +0.018 \\
Ionosphere & Cls. & \textbf{0.943}/0.871 & 0.679/0.923 & +0.072 \\
Phoneme & Cls. & \textbf{0.856}/0.845 & 0.813/0.984 & +0.011 \\
QSAR Biodeg. & Cls. & \textbf{0.891}/0.886 & 0.414/0.929 & +0.005 \\
\bottomrule
\end{tabular}%
}
\caption{Radius-control ablation. C and U denote radius-controlled and unconstrained HYDRA. The primary metric is RMSE for regression and Accuracy for classification.}
\label{tab:radius_ablation}
\end{table}

\begin{figure}[!htbp]
    \centering
    \includegraphics[width=0.5\textwidth]{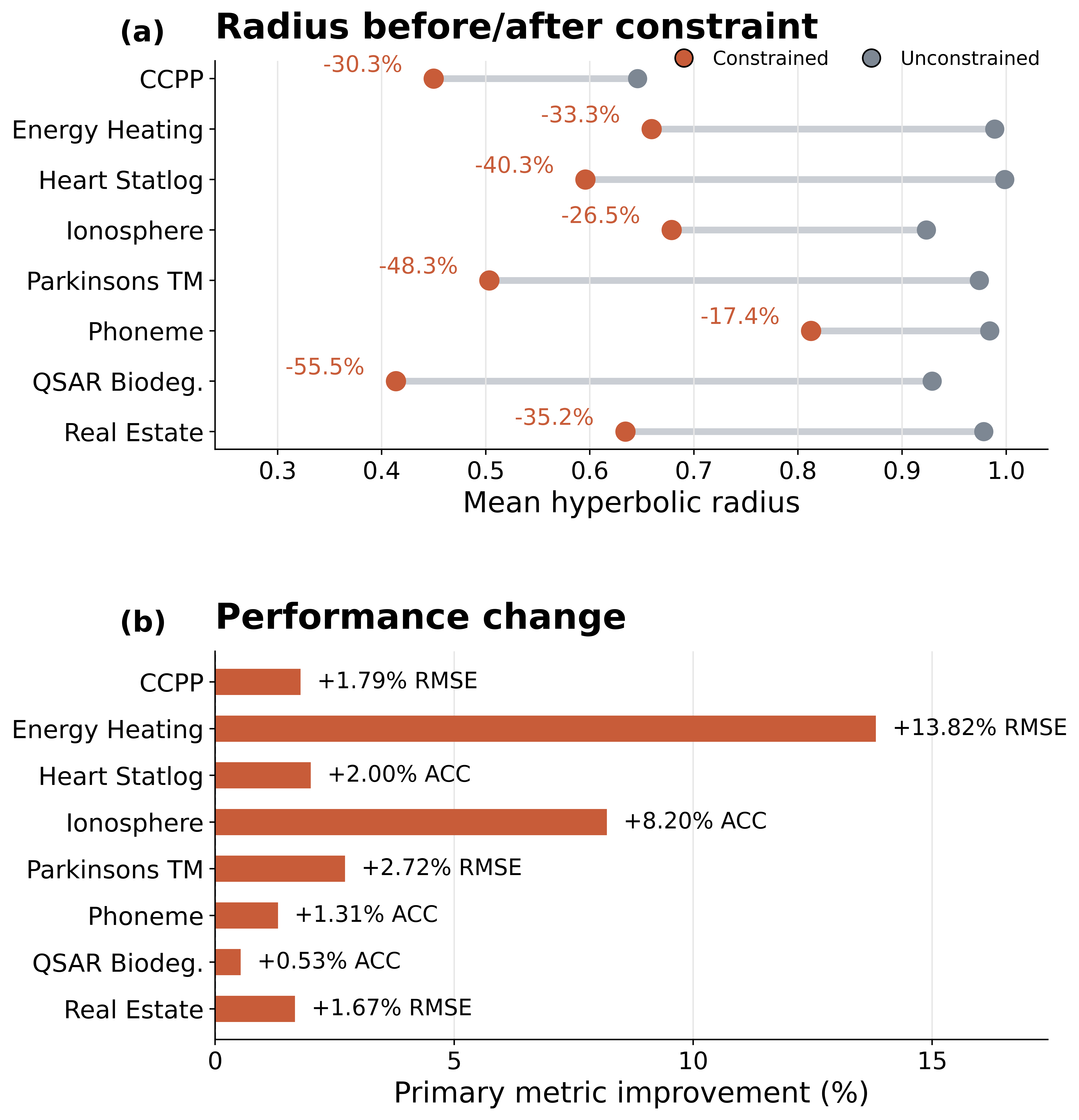}
    \caption{Radius-control ablation visualization. Constrained HYDRA reduces mean latent radius and improves the primary metric in the selected comparisons.}
    \label{fig:radius_ablation}
\end{figure}

\subsection{Ablation Studies}

The low-rank ablation asks whether prototype compression preserves performance relative to a full-rank HYDRA block. The radius ablation asks whether the hyperbolic radius budget improves optimization beyond numerical safeguarding. Tables~\ref{tab:lowrank_ablation}-\ref{tab:radius_ablation} and Figures~\ref{fig:lowrank_ablation}-\ref{fig:radius_ablation} summarize the primary results; detailed metrics are in Appendix~B.

The selected low-rank models use a mean of 46.8\% and a median of 33.8\% of the corresponding full-rank HYDRA parameters. As visualized in Figure~\ref{fig:lowrank_ablation}, compression is strongest on Heart Statlog, CCPP, Phoneme, and QSAR, while Real Estate and Ionosphere require ratios close to one, showing that the benefit is task-dependent.

\begin{figure}[!t]
    \centering
    \includegraphics[width=0.5\textwidth]{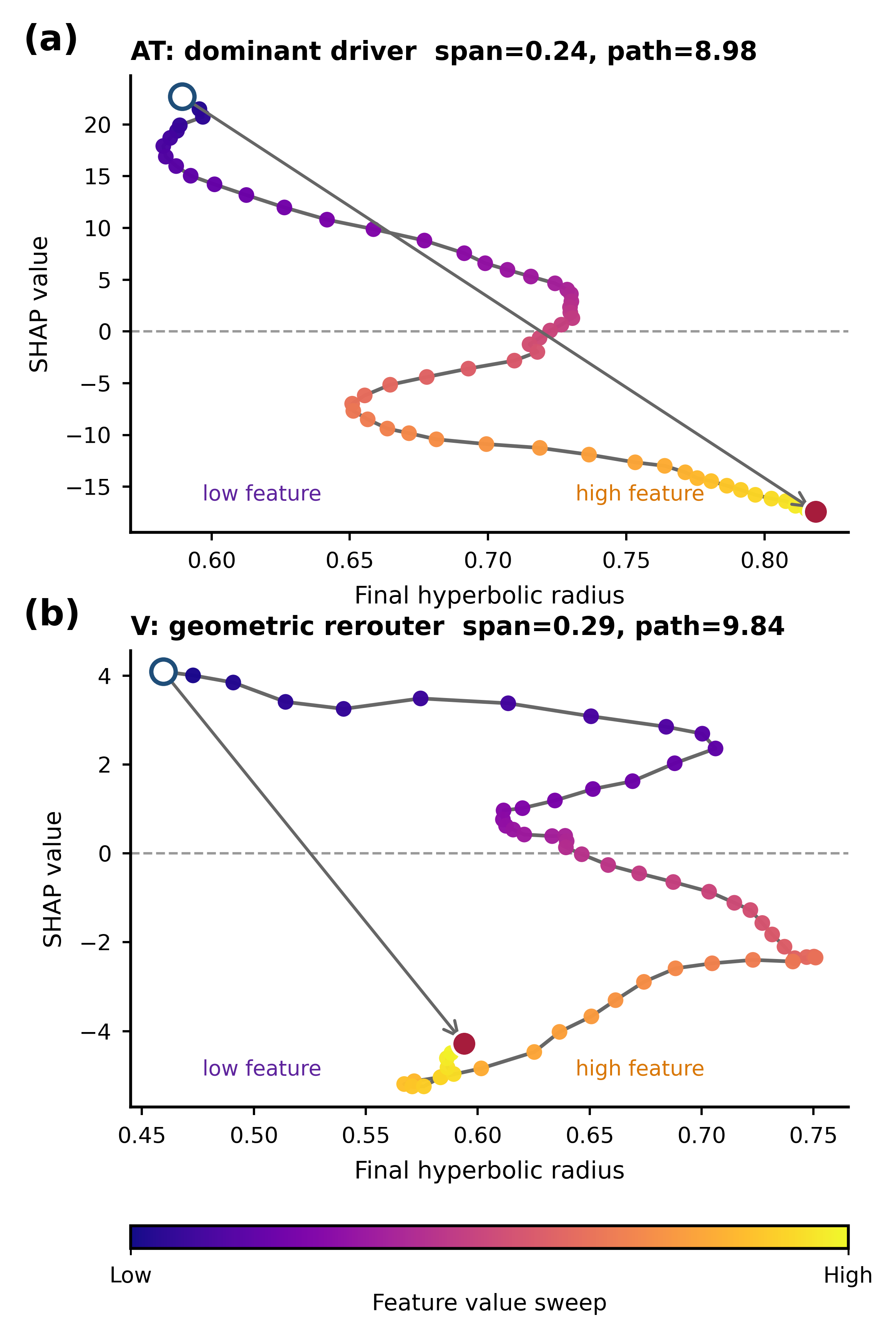}
    \caption{CCPP interpretability case study linking SHAP contribution,
    radius response, and path geometry.}
    \label{fig:interpretability_ccpp}
\end{figure}

Across all eight datasets, the constrained model has a smaller mean radius than the unconstrained model, and the primary metric improves in the selected comparisons. Figure~\ref{fig:radius_ablation} shows the same effect visually: radius control pulls representations inward while preserving or improving the primary metric. This matches the theoretical role of the radius budget: it bounds distance amplification and log-map gradients, discouraging near-boundary shortcuts.

Without radius control, the model may exploit the outer region of the Poincaré ball, where even small Euclidean perturbations correspond to large hyperbolic distances. Such movement can help separate samples, but it can also create an unstable shortcut in which the model reduces the loss by pushing representations outward instead of learning a smooth tangent-space functional response. The constrained variant limits this behavior. The simultaneous reduction in mean radius and improvement in the primary metric suggests that radius control changes the learned representation, not only the numerical range of the hidden state.

\subsection{Interpretability Analysis}

HYDRA's interpretability comes from the geometry of its hidden representation. For a controlled feature sweep, we record the final hyperbolic radius, trajectory shape, and path length of the hidden state. These quantities describe how the model reorganizes its internal representation as one input variable is changed. At the same time, we have introduced SHAP (SHapley Additive exPlanations) values as a reference to demonstrate the interpretability of the model. SHAP is a game-theoretic interpretability method that quantifies the marginal contribution of each input feature to a model prediction. By decomposing predictions into feature-level positive or negative effects, SHAP provides an intuitive explanation of complex model behavior.

For a sweep $t\mapsto(\mathbf{x}_{-j},t)$ with final representation $\gamma_j(t)$, the path length
\begin{equation}
    \mathcal{L}_j=\int\|\gamma_j'(t)\|_{\mathcal{B}_c}\,dt,\quad
\|\gamma_j'(t)\|_{\mathcal{B}_c}
=\lambda_{\gamma_j(t)}^c\|\gamma_j'(t)\|_2  
\end{equation}

weights Euclidean displacement by the local conformal factor. Large radius changes or long paths therefore indicate stronger internal reorganization.

We used the CCPP dataset as a case study to examine whether HYDRA’s latent geometry provides physically meaningful interpretability. The prediction target is net hourly electrical power output. Ambient temperature (AT), ambient pressure (AP), and relative humidity (RH) mainly describe gas-turbine operating conditions, whereas exhaust vacuum (V) reflects the steam-turbine side of the combined-cycle process. As shown in Figure~\ref{fig:interpretability_ccpp}, the AT sweep produced the clearest radius-output relation. The trajectory moved from a small-radius, positive-SHAP regime at low AT to a large-radius, negative-SHAP regime at high AT, and the one-feature PDP reduced predicted power by 37.36 MW. This agrees with the physical expectation that hotter intake air lowers air density and reduces the available mass flow through the gas turbine.

The V sweep also showed a negative output direction, but with a longer and more tortuous hyperbolic path. This pattern is consistent with a coupled steam-side condition that changes the internal representation, rather than acting as a simple direct driver. The corresponding AP and RH sweeps, reported in the Appendix, induced weaker geometric responses, consistent with their secondary roles among the operating variables. Overall, these observations support hyperbolic radius and path geometry as HYDRA-specific interpretability diagnostics. The Appendix further repeats the same visualization protocol on additional datasets to assess whether the CCPP pattern generalizes beyond this case.

\section{Conclusion}
We introduced HYDRA as a compact extension of KAN that couples tangent-space spline computation with hyperbolic latent representations. Across eight tabular benchmarks, HYDRA achieved competitive or superior predictive performance while reducing trainable parameters by 34.9\% relative to Euclidean KAN and by 37.1\% relative to MLP on average. Ablation studies showed that the low-rank prototype block retained predictive performance, and that radius control reduced near-boundary saturation in the  Poincaré ball. The CCPP interpretability case study further showed that hyperbolic radius and latent trajectories can provide HYDRA-specific diagnostic signals that are consistent with physical expectations and SHAP trends. Overall, these results indicate that hyperbolic representation geometry can support parameter-efficient KAN-style function learning while offering an inspectable latent-space view of model behavior.

% Note: \bibliographystyle{aaai2026} is automatically set by aaai2026.sty
% Do not add \bibliographystyle{aaai2026} here as it will cause "Illegal, another \bibstyle command" error
% \clearpage
\bibliography{aaai2027}

\onecolumn
\appendix

\begingroup
\small
\section{Appendix A: Universal Approximation of HYDRA}
\label{app:universal}

\paragraph{Setting.}
Let $K\subset\mathbb{R}^{p}$ be compact and
$f^\star\in C(K,\mathbb{R}^{m})$. Define
\[
\|f\|_{\infty,K}:=\sup_{\mathbf{x}\in K}\|f(\mathbf{x})\|_2 .
\]
For $0<r_{\max}<1$, let
\[
\mathcal{B}^{d}_{c,r_{\max}}
=
\{\mathbf{h}\in\mathbb{R}^{d}:\sqrt{c}\|\mathbf{h}\|_2\le r_{\max}\},
\qquad
\mathcal{T}^{d}_{c,r_{\max}}
=
\log_0^c(\mathcal{B}^{d}_{c,r_{\max}}).
\]
On these sets,
\[
\exp_0^c:\mathcal{T}^{d}_{c,r_{\max}}\rightarrow
\mathcal{B}^{d}_{c,r_{\max}},
\qquad
\log_0^c:\mathcal{B}^{d}_{c,r_{\max}}\rightarrow
\mathcal{T}^{d}_{c,r_{\max}}
\]
are continuous inverses.

\paragraph{Proposition.}
For every $\varepsilon>0$, there exist finite $d,L,K_s,r$ and HYDRA parameters
$\theta$ such that
\[
\|f_{\theta}^{\mathrm{HYDRA}}-f^\star\|_{\infty,K}<\varepsilon .
\]
Equivalently,
\[
\overline{\mathcal{F}_{\mathrm{HYDRA}}}^{\|\cdot\|_{\infty,K}}
=
C(K,\mathbb{R}^{m}),
\]
provided width, depth, spline resolution, and prototype rank are allowed to
increase.

\paragraph{Proof.}
Let $\varepsilon>0$. By spline approximation and Stone-Weierstrass
\cite{rudin1976principles}, there exists a finite Euclidean spline/KAN network
$g_{\phi}:K\rightarrow\mathbb{R}^{m}$ such that
\[
\|g_{\phi}-f^\star\|_{\infty,K}<\varepsilon/3 .
\tag{A1}
\]
Write one Euclidean KAN residual block as
\[
\mathbf{z}_{l+1}
=
\mathbf{z}_{l}
+A_{l}\Phi_{l}(B_{l}\mathbf{z}_{l}),
\tag{A2}
\]
where $\Phi_{l}$ is a coordinate-wise spline operator. A corresponding HYDRA
block is
\[
\begin{aligned}
\mathbf{h}_{l} &= \exp_0^c(\mathbf{z}_{l}),\\
\tilde{\mathbf{z}}_{l+1}
&=
\log_0^c(\mathbf{h}_{l})
+W_{\uparrow,l}\Phi_{l}
(W_{\downarrow,l}\log_0^c(\mathbf{h}_{l})),\\
\mathbf{h}_{l+1}
&=
\Pi_{r_{\max}}\!\left(\exp_0^c(\tilde{\mathbf{z}}_{l+1})\right).
\end{aligned}
\tag{A3}
\]
Choose $r\ge d$ and set
\[
W_{\downarrow,l}
=
\begin{bmatrix}I_d\\0\end{bmatrix},
\qquad
W_{\uparrow,l}
=
\begin{bmatrix}A_{l} & 0\end{bmatrix}.
\tag{A4}
\]
Then
\[
W_{\uparrow,l}\Phi_{l}(W_{\downarrow,l}\mathbf{z})
=
A_{l}\Phi_{l}(\mathbf{z}),
\tag{A5}
\]
so the low-rank prototype block contains the full tangent-space KAN update
whenever $r\ge d$.

Scale the Euclidean realization by $\alpha>0$:
\[
\mathbf{z}_{l}^{(\alpha)}=\alpha \mathbf{z}_{l},
\qquad
\alpha K \subset \mathcal{T}^{d}_{c,r_{\max}} .
\tag{A6}
\]
For $\|\mathbf{z}\|\le R_{\alpha}$ with $R_{\alpha}\rightarrow 0$,
\[
\exp_0^c(\mathbf{z})
=
\mathbf{z}+O(\|\mathbf{z}\|^3),
\qquad
\log_0^c(\mathbf{h})
=
\mathbf{h}+O(\|\mathbf{h}\|^3).
\tag{A7}
\]
Hence, for each block,
\[
\sup_{\mathbf{x}\in K}
\|\mathcal{H}_{l}^{(\alpha)}(\mathbf{x})
-\mathcal{G}_{l}^{(\alpha)}(\mathbf{x})\|_2
\le C_{l}R_{\alpha}^{3}.
\tag{A8}
\]
For finite depth $L$,
\[
\sup_{\mathbf{x}\in K}
\|H_{\theta}^{(\alpha)}(\mathbf{x})-G_{\phi}^{(\alpha)}(\mathbf{x})\|_2
\le
\sum_{l=1}^{L} C_{l}R_{\alpha}^{3}
<\varepsilon/3 .
\tag{A9}
\]
Let
\[
f_{\theta}^{\mathrm{HYDRA}}(\mathbf{x})
=
\alpha^{-1}Q H_{\theta}^{(\alpha)}(\mathbf{x}),
\qquad
g_{\phi}(\mathbf{x})
=
\alpha^{-1}Q G_{\phi}^{(\alpha)}(\mathbf{x}),
\tag{A10}
\]
and choose $\alpha$ so that
\[
\|f_{\theta}^{\mathrm{HYDRA}}-g_{\phi}\|_{\infty,K}
<\varepsilon/3 .
\tag{A11}
\]
Combining (A1) and (A11),
\[
\begin{aligned}
\|f_{\theta}^{\mathrm{HYDRA}}-f^\star\|_{\infty,K}
&\le
\|f_{\theta}^{\mathrm{HYDRA}}-g_{\phi}\|_{\infty,K}
+\|g_{\phi}-f^\star\|_{\infty,K}\\
&<2\varepsilon/3<\varepsilon .
\end{aligned}
\tag{A12}
\]
Thus $\mathcal{F}_{\mathrm{HYDRA}}$ is dense in $C(K,\mathbb{R}^{m})$.

\paragraph{Rank condition.}
\[
r\ge d
\quad\Longrightarrow\quad
\operatorname{rank}
\left(
W_{\uparrow}J_{\Phi}(W_{\downarrow}\mathbf{z})W_{\downarrow}
\right)
\le d ,
\tag{A13}
\]
and the full tangent update can be recovered. Fixed $r<d$ instead imposes
\[
\operatorname{rank}(J_{\mathrm{HYDRA}}(\mathbf{z}))\le r,
\tag{A14}
\]
so low-rank HYDRA is a parameter-efficient subfamily of the universal class.

\endgroup

\clearpage

\section{Appendix B: Supplementary Experimental Results}
\label{app:supp}

This appendix reports secondary metrics and compact ablation details for the final benchmark datasets. The main paper uses RMSE for regression and Accuracy for classification as primary metrics. Here, regression rows additionally report MAE and $R^2$, where smaller MAE and larger $R^2$ are better. Classification rows report F1 and ROC-AUC, where larger values are better.
HYDRA is the proposed model; GAMI denotes GAMI-Net; HGCN and HNN are hyperbolic neural baselines; KAN, MLP, NAM, and NODE denote the Euclidean baseline families. Bold marks the best value in a row and underlining marks the second-best value.

For ablations, $r$ is the selected prototype rank, and Param ratio is the low-rank/full-rank trainable-parameter ratio. F/L denotes full-rank versus selected low-rank HYDRA. C/U denotes radius-controlled versus unconstrained HYDRA. Aux. lists the secondary metrics in the same order as above. Radius P95 is the 95th percentile final radius; lower values indicate less near-boundary saturation.

\begin{center}
\scriptsize
\renewcommand{\arraystretch}{1.08}
\setlength{\tabcolsep}{4pt}
\begin{tabular*}{\textwidth}{@{\extracolsep{\fill}}llcccccccc@{}}
\toprule
Task & Dataset & HYDRA & GAMI & HGCN & HNN & KAN & MLP & NAM & NODE \\
\midrule
\multicolumn{10}{@{}l}{\textit{Regression: MAE / $R^2$}} \\
Reg. & CCPP & \textbf{2.700}/\textbf{0.955} & 2.991/0.949 & 3.308/0.940 & 3.202/0.944 & \underline{2.746}/\underline{0.954} & 2.874/0.951 & 3.236/0.942 & 3.223/0.943 \\
Reg. & Energy & \textbf{0.524}/\textbf{0.995} & 1.101/0.980 & 1.743/0.941 & 1.635/0.948 & \underline{0.533}/\underline{0.995} & 1.062/0.982 & 1.924/0.910 & 0.859/0.987 \\
Reg. & Parkinsons & \textbf{2.479}/\textbf{0.892} & 9.563/-0.140 & 5.954/0.481 & 4.997/0.617 & \underline{3.211}/\underline{0.830} & 3.708/0.756 & 7.829/0.219 & 5.786/0.546 \\
Reg. & Real Estate & \textbf{4.829}/\textbf{0.748} & 4.994/0.722 & 5.743/0.672 & 5.524/0.686 & 5.792/0.646 & \underline{4.875}/\underline{0.744} & 7.242/0.564 & 5.286/0.710 \\
\midrule
\multicolumn{10}{@{}l}{\textit{Classification: F1 / ROC-AUC}} \\
Cls. & Heart & \textbf{0.933}/\textbf{0.970} & 0.909/0.949 & \underline{0.930}/0.944 & \underline{0.930}/0.954 & 0.780/0.924 & 0.875/\underline{0.967} & 0.750/0.867 & 0.844/0.926 \\
Cls. & Ionosphere & \textbf{0.979}/\textbf{0.974} & \underline{0.959}/0.962 & 0.900/0.949 & 0.922/\underline{0.967} & 0.887/0.938 & 0.920/0.953 & 0.863/0.728 & 0.928/0.929 \\
Cls. & Phoneme & \textbf{0.807}/\textbf{0.943} & 0.701/0.911 & 0.702/0.903 & 0.717/0.902 & \underline{0.772}/\underline{0.924} & 0.737/0.913 & 0.643/0.808 & 0.683/0.893 \\
Cls. & QSAR & \textbf{0.857}/\textbf{0.931} & 0.797/0.905 & 0.822/0.923 & \underline{0.841}/0.919 & 0.757/0.891 & 0.830/\underline{0.927} & 0.707/0.846 & 0.783/0.922 \\
\bottomrule
\end{tabular*}
\captionof{table}{Secondary metrics on the final benchmark datasets.}
\label{tab:appendix_secondary_metrics}
\label{tab:appendix_regression_metrics}
\label{tab:appendix_classification_metrics}
\end{center}

\begin{center}
\scriptsize
\renewcommand{\arraystretch}{1.08}
\setlength{\tabcolsep}{4pt}
\begin{tabular*}{\textwidth}{@{\extracolsep{\fill}}llrrccccc@{}}
\toprule
Task & Dataset & $r$ & Param ratio & Primary F/L & Aux. F/L & Primary C/U & Aux. C/U & Radius P95 C/U \\
\midrule
Reg. & CCPP & 3 & 0.13 & 3.947/\textbf{3.919} & 3.072/\textbf{3.051}; 0.946/\textbf{0.947} & \textbf{3.624}/3.690 & \textbf{2.712}/2.757; \textbf{0.955}/0.953 & \textbf{0.468}/0.763 \\
Reg. & Energy & 3 & 0.67 & 2.067/\textbf{1.959} & 1.580/\textbf{1.536}; 0.957/\textbf{0.962} & \textbf{1.193}/1.384 & \textbf{0.878}/1.032; \textbf{0.986}/0.981 & \textbf{0.754}/0.999 \\
Reg. & Parkinsons & 3 & 0.40 & 6.342/\textbf{6.037} & 4.737/\textbf{4.430}; 0.651/\textbf{0.684} & \textbf{5.723}/5.883 & \textbf{4.043}/4.284; \textbf{0.716}/0.700 & \textbf{0.698}/1.000 \\
Reg. & Real Estate & 6 & 0.99 & 7.273/\textbf{6.769} & 4.981/\textbf{4.829}; 0.709/\textbf{0.748} & \textbf{6.769}/6.884 & \textbf{4.829}/4.922; \textbf{0.748}/0.739 & \textbf{0.753}/1.000 \\
\midrule
Cls. & Heart & 1 & 0.09 & 0.889/\textbf{0.926} & 0.875/\textbf{0.909}; \textbf{0.964}/0.960 & \textbf{0.944}/0.926 & \textbf{0.930}/0.909; 0.970/\textbf{0.971} & \textbf{0.665}/1.000 \\
Cls. & Ionosphere & 7 & 0.95 & 0.871/\textbf{0.943} & 0.907/\textbf{0.958}; \textbf{0.957}/\textbf{0.957} & \textbf{0.943}/0.871 & \textbf{0.958}/0.909; \textbf{0.966}/0.931 & \textbf{0.799}/1.000 \\
Cls. & Phoneme & 4 & 0.25 & 0.836/\textbf{0.840} & 0.684/\textbf{0.747}; \textbf{0.914}/\textbf{0.914} & \textbf{0.856}/0.845 & \textbf{0.747}/0.722; \textbf{0.917}/0.915 & \textbf{0.884}/1.000 \\
Cls. & QSAR & 1 & 0.28 & 0.877/\textbf{0.882} & 0.824/\textbf{0.828}; \textbf{0.923}/0.920 & \textbf{0.891}/0.886 & \textbf{0.844}/0.836; \textbf{0.928}/0.924 & \textbf{0.681}/1.000 \\
\bottomrule
\end{tabular*}
\captionof{table}{Combined low-rank and radius-control ablations.}
\label{tab:appendix_ablation_summary}
\label{tab:appendix_lowrank_ablation}
\label{tab:appendix_radius_ablation}
\label{tab:appendix_lowrank_regression}
\label{tab:appendix_lowrank_classification}
\label{tab:appendix_radius_regression}
\label{tab:appendix_radius_classification}
\end{center}

\section{Appendix C: Additional Interpretability Cases}
\label{app:interpretability}

Figure~\ref{fig:appendix_interpretability_other_datasets} extends the CCPP case study to additional datasets. Each panel links the SHAP contribution of a controlled feature sweep to the final hyperbolic radius reached by HYDRA. These cases are used as output-side references: HYDRA's own diagnostics are the radius, trajectory, and latent-path changes, while SHAP summarizes the observed prediction contribution.

\begin{center}
\includegraphics[width=\textwidth]{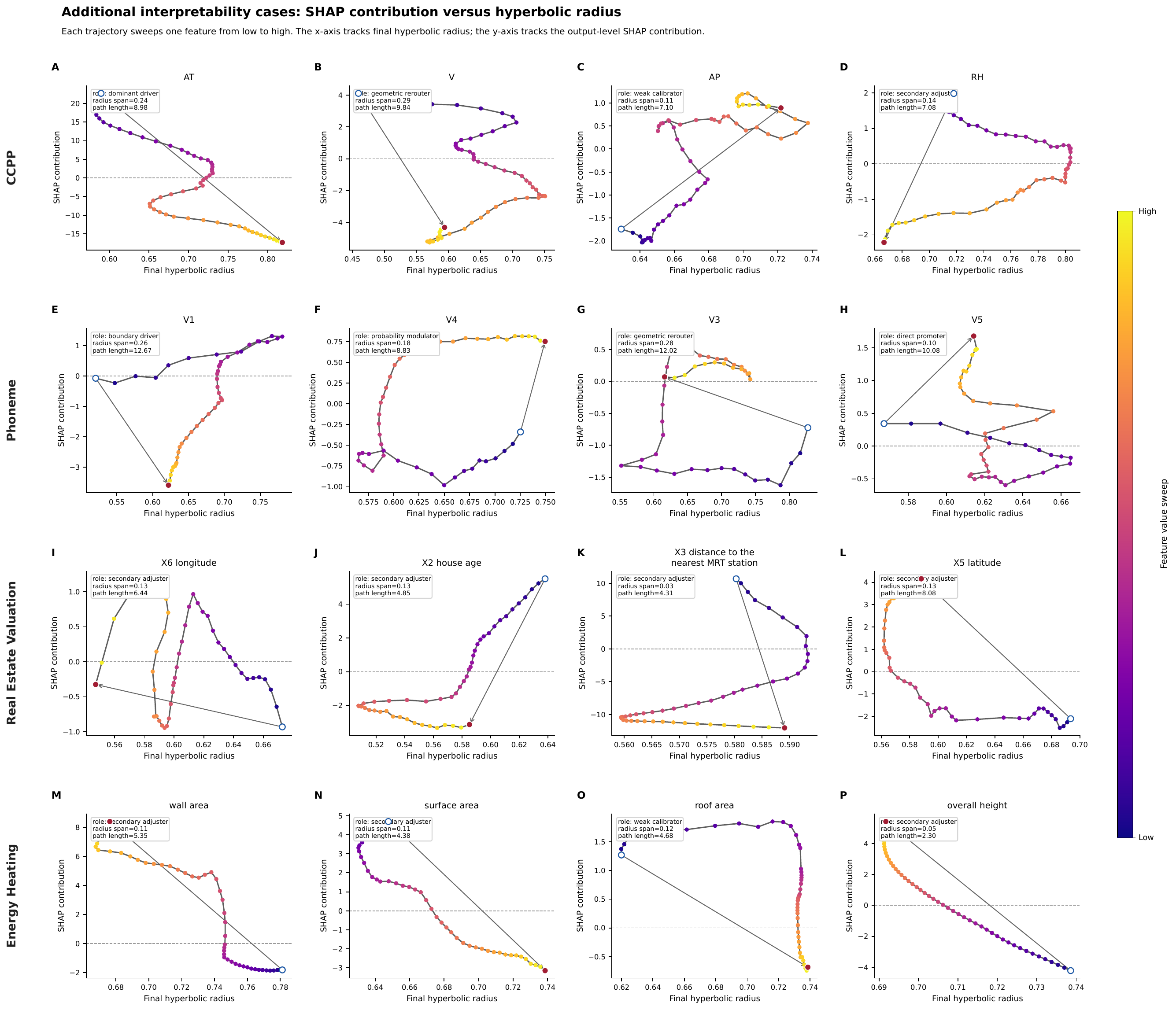}
\captionof{figure}{Additional interpretability cases beyond CCPP. Each panel traces how sweeping a representative feature changes both final hyperbolic radius and SHAP contribution.}
\label{fig:appendix_interpretability_other_datasets}
\end{center}

\section{Appendix D: Training Hyperparameters}
\label{app:hyperparameters}

This appendix reports the training hyperparameters used for the main comparison in Table~\ref{tab:main_results_with_params}. All runs use seed 42. Continuous features are normalized using training-set statistics. Regression targets are optimized in normalized space and mapped back to the original scale for reporting. Classification runs optimize binary cross-entropy with logits.

For HYDRA, Ep. is the number of training epochs and B is the mini-batch size.  $w$ is hidden width, $K$ is the number of spline knots, and $r$ is the low-rank prototype rank. LR and WD are the optimizer learning rate and weight decay. Drop. is dropout probability. Rad. is reported as radius-loss weight / target-radius ratio. Pos. is the positive-class loss weight; regression rows use 1.00 because no class reweighting is applied.

For compact baseline tables, each cell begins with Ep./B and then reports the architecture. KAN cells use $w,K$; MLP cells use $w$; HGCN and HNN cells use $w,K$ followed by Rad. GAMI-Net cells report main subnet width, interaction subnet width, $n_{\rm int}$, and main-stage/fine-tuning LR. NAM cells report shallow feature-network width, hidden feature-network width, and
LR/WD; ``none'' means no additional hidden feature-network layer. NODE-GAM cells report trees, tree layers, tree depth, column subsampling ratio, and LR/$L_2$.

\begin{center}
\scriptsize
\renewcommand{\arraystretch}{1.08}
\setlength{\tabcolsep}{4pt}
\begin{tabular*}{\textwidth}{@{\extracolsep{\fill}}lrrrrrccccc@{}}
\toprule
Dataset & Ep. & B & $w$ & $K$ & $r$ & LR & WD & Drop. & Rad. & Pos. \\
\midrule
CCPP & 120 & 256 & 43 & 6 & 16 & $8.5{\times}10^{-4}$ & 0 & 0 & 0/0.88 & 1.00 \\
Energy Heating & 80 & 256 & 20 & 6 & 8 & $1.5{\times}10^{-3}$ & $10^{-4}$ & 0 & 0/0.90 & 1.00 \\
Parkinsons TM & 220 & 256 & 13 & 4 & 12 & $1.8{\times}10^{-3}$ & $10^{-4}$ & 0 & 0.03/0.88 & 1.00 \\
Real Estate & 140 & 128 & 28 & 6 & 6 & $9.0{\times}10^{-4}$ & $10^{-4}$ & 0 & 0/0.90 & 1.00 \\
Heart Statlog & 80 & 256 & 47 & 4 & 2 & $7.0{\times}10^{-4}$ & $5{\times}10^{-5}$ & 0.02 & 0/0.90 & 1.20 \\
Ionosphere & 120 & 256 & 10 & 4 & 8 & $1.0{\times}10^{-3}$ & $10^{-4}$ & 0 & 0/0.90 & 1.00 \\
Phoneme & 120 & 128 & 15 & 6 & 8 & $8.0{\times}10^{-4}$ & $5{\times}10^{-5}$ & 0.02 & 0.03/0.90 & 0.95 \\
QSAR Biodeg. & 140 & 256 & 27 & 6 & 2 & $6.5{\times}10^{-4}$ & $10^{-4}$ & 0.015 & 0/0.90 & 1.00 \\
\bottomrule
\end{tabular*}
\captionof{table}{Final HYDRA hyperparameters on the eight benchmark datasets.}
\label{tab:appendix_hkan_hparams}
\end{center}

\begin{center}
\scriptsize
\renewcommand{\arraystretch}{1.08}
\setlength{\tabcolsep}{5pt}
\begin{tabular*}{\textwidth}{@{\extracolsep{\fill}}p{0.13\textwidth}p{0.16\textwidth}p{0.11\textwidth}p{0.20\textwidth}p{0.20\textwidth}@{}}
\toprule
Dataset & KAN & MLP & HGCN & HNN \\
\midrule
CCPP & 120/256; 51,12 & 120/256; 56 & 120/256; 79,4; 0/0.90 & 120/256; 41,4; 0/0.90 \\
Energy Heating & 80/256; 19,12 & 80/256; 20 & 80/256; 25,4; 0/0.90 & 80/256; 26,4; 0/0.90 \\
Parkinsons TM & 100/512; 39,6 & 100/512; 28 & 100/512; 23,4; 0/0.90 & 100/512; 24,4; 0/0.90 \\
Real Estate & 80/256; 38,6 & 80/256; 36 & 80/256; 36,4; 0/0.90 & 80/256; 37,4; 0/0.90 \\
Heart Statlog & 80/256; 9,12 & 80/256; 82 & 80/256; 29,4; 0/0.90 & 80/256; 29,4; 0/0.90 \\
Ionosphere & 80/256; 13,8 & 80/256; 47 & 80/256; 21,4; 0/0.90 & 80/256; 27,4; 0/0.90 \\
Phoneme & --/--; 23,8 & --/--; 37 & --/--; 37,4; 0/0.90 & --/--; 38,4; 0/0.90 \\
QSAR Biodeg. & --/--; 27,4 & --/--; 24 & --/--; 30,4; 0/0.90 & --/--; 31,4; 0/0.90 \\
\bottomrule
\end{tabular*}
\captionof{table}{Internal neural baseline hyperparameters.}
\label{tab:appendix_internal_baseline_hparams}
\label{tab:appendix_euclidean_baseline_hparams}
\label{tab:appendix_hyperbolic_baseline_hparams}
\end{center}

\begin{center}
\scriptsize
\renewcommand{\arraystretch}{1.08}
\setlength{\tabcolsep}{5pt}
\begin{tabular*}{\textwidth}{@{\extracolsep{\fill}}p{0.14\textwidth}p{0.23\textwidth}p{0.22\textwidth}p{0.23\textwidth}@{}}
\toprule
Dataset & GAMI-Net & NAM & NODE-GAM \\
\midrule
CCPP & 100/256; 12; 14--26; 10; $10^{-3}/10^{-4}$ & 140/256; 21; 20--62; $5{\times}10^{-4}$/0 & 100/256; 160,1,6,0.5; $10^{-2}/10^{-5}$ \\
Energy Heating & 80/256; 12; 12--12; 4; $10^{-3}/10^{-4}$ & 140/256; 4; 32; $5{\times}10^{-4}$/0 & 80/256; 143,1,3,1.0; $10^{-2}/10^{-5}$ \\
Parkinsons TM & 100/256; 12; 12--12; 8; $10^{-3}/10^{-4}$ & 140/256; 8; 12; $5{\times}10^{-4}$/$10^{-6}$ & 100/256; 64,1,6,0.5; $10^{-2}/10^{-5}$ \\
Real Estate & 80/256; 12; 16--8; 8; $10^{-3}/10^{-4}$ & 116/256; 8; 16--8; $5{\times}10^{-4}$/0 & 80/256; 128,1,2,0.5; $10^{-2}/10^{-5}$ \\
Heart Statlog & 80/256; 32; 8--8; 4; $7{\times}10^{-4}/7{\times}10^{-5}$ & 140/256; 5; 15; $5{\times}10^{-4}$/0 & 80/256; 27,2,2,0.5; $10^{-2}/10^{-5}$ \\
Ionosphere & 80/256; 14; 7; 7; $10^{-3}/10^{-4}$ & 140/256; 16; none; $5{\times}10^{-4}$/0 & 80/256; 48,1,2,0.5; $10^{-2}/10^{-5}$ \\
Phoneme & 80/256; 12; 10--14; 8; $10^{-3}/10^{-4}$ & 140/256; 12; 16--8; $5{\times}10^{-4}$/0 & 80/256; 64,1,6,0.5; $10^{-2}/10^{-5}$ \\
QSAR Biodeg. & 80/256; 8; 8--18; 4; $10^{-3}/10^{-4}$ & 140/256; 24; none; $5{\times}10^{-4}$/0 & 80/256; 64,1,4,0.5; $10^{-2}/10^{-5}$ \\
\bottomrule
\end{tabular*}
\captionof{table}{External GAM baseline hyperparameters.}
\label{tab:appendix_external_gam_hparams}
\label{tab:appendix_gaminet_hparams}
\label{tab:appendix_nam_hparams}
\label{tab:appendix_nodegam_hparams}
\end{center}

\section{Appendix E: Limitations}
This work has several limitations. First, our empirical evaluation focuses on standard tabular benchmarks, leaving the behavior of HYDRA on high-dimensional unstructured data (e.g., images, text, and multimodal inputs) for future investigation. Second, the proposed multi-feature analysis should be interpreted as a post hoc diagnostic of the learned model rather than a causal explanation. While it reveals that output non-additivity can coincide with hyperbolic latent reorganization, it does not establish causal or physical interactions among input variables. Finally, although we compare HYDRA with additional parameter-efficient KAN variants, its prototype rank and radius remain validation-dependent hyperparameters. Developing adaptive strategies for rank selection and radius scheduling could further improve robustness across datasets while reducing the need for manual tuning.

\end{document}